\documentclass[letterpaper, 10 pt, conference]{ieeeconf}  % Comment this line out if you need a4paper

\IEEEoverridecommandlockouts                              % This command is only needed if 
\usepackage[dvipsnames]{xcolor}
\usepackage{amsmath}
\usepackage{amsfonts}
\usepackage{graphicx}
\usepackage{subcaption}
\usepackage{cite}
\let\labelindent\relax
\usepackage{enumitem}
\usepackage{booktabs}
\usepackage{multirow}
\usepackage[hyphens]{url}
\usepackage[hidelinks]{hyperref}
\usepackage{soul,color}

\title{\LARGE \bf
Visual Object Recognition in Indoor Environments Using Topologically Persistent Features}

\author{Ekta U. Samani$^{1}$, Xingjian Yang$^{1}$, and Ashis G. Banerjee$^{2}$% <-this % stops a space
\thanks{*This work has been accepted for publication in the IEEE Robotics and Automation Letters, doi: \href{https://ieeexplore.ieee.org/document/9496193}{10.1109/LRA.2021.3099460.}
This work was supported in part by the 2019 Amazon Research Award.}% <-this % stops a space
\thanks{$^{1}$E. Samani and X. Yang are with the Department of Mechanical Engineering, University of Washington, Seattle, WA 98195, USA,
        {\tt\small ektas@uw.edu,yxj1995@uw.edu}}%

\thanks{$^{2}$A. G. Banerjee is with the Department of Industrial \& Systems Engineering and the Department of Mechanical Engineering, University of Washington, Seattle, WA 98195, USA,
        {\tt\small ashisb@uw.edu}}%
}

\begin{document}

\maketitle
\thispagestyle{empty}
\pagestyle{empty}

%%%%%%%%%%%%%%%%%%%%%%%%%%%%%%%%%%%%%%%%%%%%%%%%%%%%%%%%%%%%%%%%%%%%%%%%%%%%%%%%
\begin{abstract}

Object recognition in unseen indoor environments remains a challenging problem for visual perception of mobile robots. In this letter, we propose the use of topologically persistent features, which rely on the objects’ shape information, to address this challenge. In particular, we extract two kinds of features, namely, sparse persistence image (PI) and amplitude, by applying persistent homology to multi-directional height function-based filtrations of the cubical complexes representing the object segmentation maps. The features are then used to train a fully connected network for recognition.
For performance evaluation, in addition to a widely used shape dataset and a benchmark indoor scenes dataset, we collect a new dataset, comprising scene images from two different environments, namely, a living room and a mock warehouse. 
% The scenes in both the environments include up to five different objects that are chosen from a given set of fourteen objects. The objects have varying poses and arrangements, and are imaged under different illumination conditions and camera poses.
The scenes are captured using varying camera poses under different illumination conditions and include up to five different objects from a given set of fourteen objects. On the benchmark indoor scenes dataset, sparse PI features show better recognition performance in unseen environments than the features learned using the widely used ResNetV2-56 and EfficientNet-B4 models. Further, they provide slightly higher recall and accuracy values than Faster R-CNN, an end-to-end object detection method, and its state-of-the-art variant, Domain Adaptive Faster R-CNN. 
% in unseen environments. of both the datasets.
The performance of our methods also remains relatively unchanged from the training environment (living room) to the unseen environment (mock warehouse) in the new dataset. In contrast, the performance of the object detection methods drops substantially. We also implement the proposed method on a real-world robot to demonstrate its usefulness.

%%Recognition using sparse PI features achieves 87\% accuracy on the shape dataset and 71\% accuracy for the living room environment in the more challenging indoor scenes dataset.


\end{abstract}

%%%%%%%%%%%%%%%%%%%%%%%%%%%%%%%%%%%%%%%%%%%%%%%%%%%%%%%%%%%%%%%%%%%%%%%%%%%%%%%%

\section{Introduction}

%story
%object recognition is a key capability for robot perception
%deep learning methods achieve tremendous success but known to fail in new environments
%Unsuitable for long term autonomy
%for usability across different environments features used for recognition must be independent of illumination, context, color; possible approach is to use topology
%explain topology, applications of topology
Perception is one of the core capabilities for autonomous mobile robots, and vision plays a key role in developing a semantic-level understanding of the robot's environment. Object recognition and localization, often referred together as detection, therefore, form an essential aspect of robot visual perception. Deep learning models for object recognition \cite{he2016identity,tan2019efficientnet} and object detection \cite{ren2015faster,liu2016ssd,redmon2016you} have achieved tremendous success in recognizing and detecting objects even in cluttered or crowded scenes.
% Deep learning models, such as R-CNN and its variants \cite{girschick_rcnn, girshick2015fast,ren2015faster}; and, YOLO \cite{redmon2016you} and its variants, have achieved tremendous success in detecting objects even in cluttered or crowded scenes. 
Such models, however, tend to require a large amount of training data. Therefore, a common practice is to use a model that is pre-trained on large databases with millions of images such as ImageNet, and then fine-tune the model based on scene images from the environments where the robots are expected to operate. This practice becomes cumbersome and runs into challenges for long-term robot autonomy, where the robots operate in complex and continually-changing environments for extended periods of time. Moreover, it renders the models sensitive to variations in environmental conditions such as illumination and object texture \cite{cheng2018revisiting,thys2019fooling}. Therefore, perception robustness becomes critical in such applications \cite{kunze2018artificial}.

Different human interaction-based \cite{kasaei2018towards} and human supervision-based \cite{eriksen2018learning} methods have been developed recently to address this challenge. Additionally, various domain adaptation methods have also been proposed for cross-domain object detection, i.e., detection in environments that have distributions different from that of the original training environment\cite{li2020deep}. For instance, within the Faster R-CNN framework, adversarial learning has been used for image and object instance level adaptation \cite{chen2018domain} and feature alignment \cite{he2019multi}. Saito et al. \cite{saito2019strong} use adversarial learning for strong local and weak global alignment of features in an unsupervised setting. Hsu et al. \cite{hsu2020progressive} also use adversarial learning but perform progressive adaptation through an intermediate domain, which is synthesized from the source domain images to mimic the target domain. Inoue et al. \cite{inoue2018cross} propose a two-step progressive adaptation technique %that makes use of image-level instance labels for the target domain
in a weakly-supervised setting where the detector is fine-tuned on samples generated using CycleGAN and pseudo-labeling. Similarly, Kim et al. \cite{kim2019diversify} use CycleGAN to perform domain diversification, followed by multidomain-invariant representation learning.

Alternatively, we can consider domain-invariant shape-based features in such continually-changing environments to make object recognition more robust to illumination, context, color, and texture variations (we still expect these features to be important though). Topological data analysis (TDA) focuses on extracting such shape information from high-dimensional data using algebraic and computational topology. In particular, persistent homology has been widely used to extract topologically persistent features for machine learning tasks \cite{pachauri2011topology}, especially, computer vision \cite{reininghaus2015stable,guo2018sparse,garin2019topological}. Reininghaus et al. \cite{reininghaus2015stable} propose a new stable representation that suits learning tasks, known as persistence images (PIs), and demonstrate its use for 3D shape classification/retrieval and texture recognition. Guo et al. \cite{guo2018sparse} use sparsely sampled PIs for human posture recognition and texture classification. Garin et al. \cite{garin2019topological} use persistent features from different filtration functions and representations to classify hand-written digits. Generating approximate PIs using a deep neural network has also been explored for image classification \cite{som2020pi}.

In this work, we propose the use of topologically persistent features for object recognition in indoor environments. Particularly, the key contributions of our work are as follows:

\begin{itemize}[leftmargin=*]

     \item We propose two kinds of topologically persistent features, namely, sparse PI and amplitude, for object recognition.
     
    \item We present a new dataset, the UW Indoor Scenes dataset, to evaluate the robustness of object recognition methods on unseen environments.
    \item We show that recognition using topologically persistent features is more robust to changing environments than a state-of-the-art cross-domain object detection model.
    \item We demonstrate that sparse PI features have better recognition performance than features from deep learning-based recognition methods and lead to better performance than end-to-end object detection methods (in terms of accuracy and recall) in unseen environments.
\end{itemize}
% We show that sparse PI features have better recognition performance than deep learning model-based features in an unseen environment. Additionally, we show that recognition using topologically persistent features provides greater robustness to changing environments than a state-of-the-art cross-domain object detection model. 
We also successfully implement the proposed framework on a real-world robot.

\section{Mathematical Preliminaries}
\label{math}

We begin by covering some of the mathematical preliminaries associated with TDA.

%\vspace{-1mm}
\subsection{Cubical Complexes}

For TDA, data is often represented by cubical or simplicial complexes depending on the type of data. Images can be considered as point clouds by treating every pixel as a point in $\mathbb{R}^{2}$. Such a point cloud is commonly represented using a simplicial complex. However, since images are made up of pixels, they have a natural grid-like structure to them. Therefore, they are more efficiently represented as cubical complexes in various ways \cite{robins2011theory,garin2019topological}. 

A cubical complex in $\mathbb{R}^{n}$ is a finite set of elementary cubes aligned on the grid $\mathbb{Z}^{n}$, where an elementary cube is a finite product of elementary intervals with dimension given by the number of its non-degenerate components \cite{kaczynski2006computational}. An $n$-dimensional image is a map $\mathcal{I}: I \subseteq \mathbb{Z}^{n} \longrightarrow \mathbb{R}$. A voxel is an element $v \in I$, and its value $\mathcal{I}(v)$ is the intensity. When $n=2$, the voxel is known as a pixel, and the intensity is known as the grayscale value. While there are various ways of constructing a cubical complex from an $n$-dimensional image, Garin et al. \cite{garin2019topological} adopt a method in which an $n$-cube represents a voxel, and all the adjacent lower-dimensional cubes (faces of the $n$-cube) are included. The values of the voxels $v$ are extended to all the cubes $\lambda$ in the resulting cubical complex $C$ as 
\begin{equation}
\label{buildingComplex}
\mathcal{I}^{\prime}(\lambda):=\min _{\lambda \text { is face of } v} \mathcal{I}(v).
\end{equation}
After the complexes are generated, a filtration is constructed, as described next.
%\vspace{-1mm}
\subsection{Filtration}
For any cubical (or, simplicial) complex $K$, let $K_{i}$ denote the $i$-th sublevel set of $K$. A filtration is then defined as a collection of complexes $\left\{ K_{s} \right\}_{s \in \mathbb{R}}$ such that $K_{s}$ is a subcomplex of $K_{t}$, for each $s\leq t$.

The pixel values in a grayscale image embed a natural filtration, which are used to obtain the sublevel sets of the corresponding cubical complex. The $i$-th sublevel set of the cubical complex $C$ is then obtained as
\begin{equation}
\label{sublevelSets}
C_{i}:=\left\{\lambda \in C \mid \mathcal{I}^{\prime}(\lambda) \leq i\right\},
\end{equation}
which defines the filtration as $\left\{ C_{i} \right\}_{i \in \mathcal{I}}$. For binary images, various functions known as descriptor functions (or, filtration functions) are used to construct such grayscale images \cite{garin2019topological}. Similarly, filtration functions are also defined for generating filtrations from point cloud data or mesh data \cite{reininghaus2015stable}.
%, indexed by the value of the function $\mathcal{I}$, is therefore defined 
%\vspace{-1mm}
\subsection{Persistent Homology}

Persistent homology is a common tool in TDA that studies the topological changes of the sublevel sets $C_{i}$ as $i$ increases from $(-\infty,\infty)$. During filtration, topological features, interpreted as $h$-dimensional holes, appear and disappear at different scales, referred to as their birth and death times, respectively. This information is summarized in an $h$-dimensional persistence diagram (PD). An $h$-dimensional PD is a countable multiset of points in $\mathbb{R}^{2}$. Each point $(x,y)$ represents an $h$-dimensional hole born at a time $x$ and filled at a time $y$. The diagonal of a PD is a multiset $\Delta = \left\{ (x,x) \in \mathbb{R}^{2} \vert x \in \mathbb{R}\right\}$, where every point in $\Delta$ has infinite multiplicity.

Several other stable representations of persistence can be obtained from a PD \cite{ bubenik2015statistical,adams2017persistence}. One such representation is the Persistence Image (PI), a stable and finite dimensional vector representation of persistent homology \cite{adams2017persistence}. To obtain a PI, an equivalent diagram of birth-persistence points, i.e., $(x,y-x)$, is computed. The birth-persistence points are then regarded as a sum of Dirac delta functions, which are convolved with a Gaussian kernel over a rectangular grid of evenly sampled points to compute the PI.

% The PD is first mapped to an integrable function $\rho : \mathbb{R} \longrightarrow \mathbb{R}^{2}$ known as the persistence surface, which is defined as the weighted sum of Gaussian functions centered at each point in the PD. A grid is obtained by discretizing a sub-domain of the persistence surface. Integrating the persistence surface over each grid box results in a matrix known as the PI.
%\vspace{-1mm}
\section{Methods}
\label{method}
\begin{figure*}
    \centering
    \includegraphics[width=\textwidth]{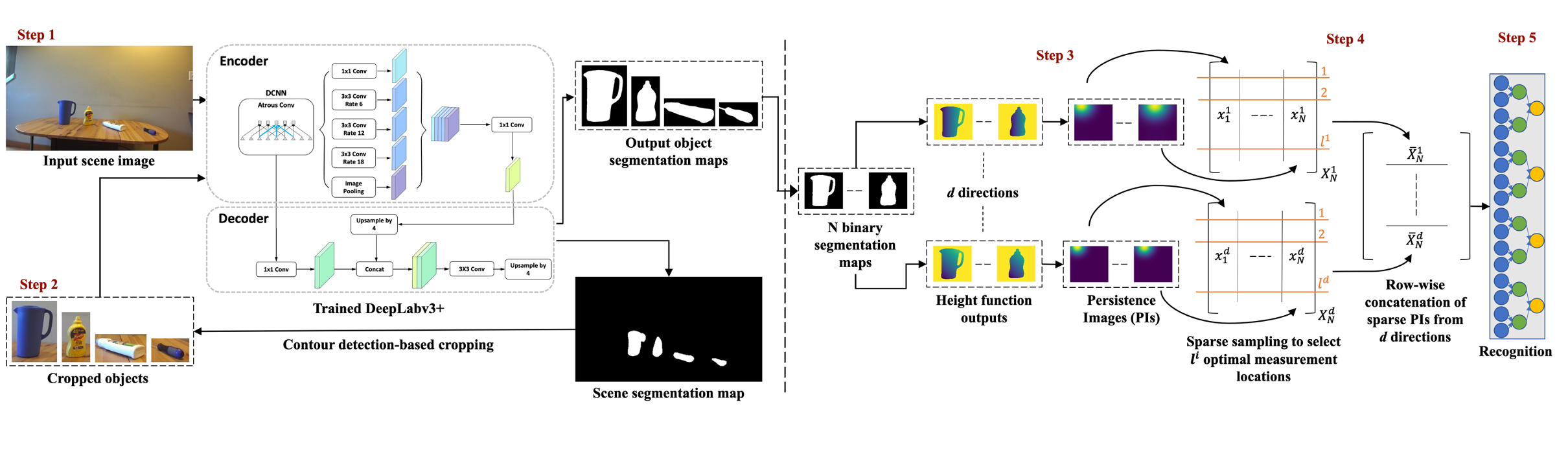}
    \caption{Proposed framework for object recognition using topologically persistent features}
    \label{pipeline}
\end{figure*}

Given an RGB scene image, our goal is to recognize all the objects in the scene, based on their shape information that are captured using topologically persistent features. We first generate segmentation maps for the objects using a deep neural network, as explained in Section \ref{maskGeneration}. We then extract topologically persistent features from the object segmentation maps, as described in \ref{featureGeneration}. These features are then fed to a fully connected network for recognition. Fig. \ref{pipeline} illustrates the proposed framework.

%tried the following way for including step numbers but discarded it

% Given an RGB scene image, our goal is to recognize all the objects in the scene, based on their shape information that are captured using topologically persistent features. Fig. \ref{pipeline} illustrates the proposed framework. We first generate segmentation maps for the objects using a deep neural network, as explained in Section \ref{maskGeneration} (Steps 1 and 2 in Fig. \ref{pipeline}). We then extract topologically persistent features from the object segmentation maps, as described in \ref{featureGeneration} (Steps 3 and 4 in Fig. \ref{pipeline}). These features are then fed to a fully connected network for recognition (Step 5 in Fig. \ref{pipeline}). 

%\vspace{-1mm}
\subsection{Object Segmentation Map Generation}
\label{maskGeneration}

To generate the object segmentation maps for an input scene image, we follow a foreground segmentation method that is similar to the one proposed in \cite{xiong2018pixel}. In particular, we use the state-of-the-art DeepLabv3+ architecture \cite{deeplabv3plus2018}, which is pre-trained on a large number of classes and is hypothesized to have a strong representation of 'objectness'. Subsequently, we train the network using pixel-level foreground annotations for a limited number of images from our datasets. 

A shape-based object recognition method relies on the objects' contours for distinguishing among multiple objects. However, the number of foreground pixels is very low as compared to the background when segmentation is performed on images taken from distances as large as two meters. Hence, it is difficult for a segmentation model to capture the minor details in the objects' shapes in a single shot. Therefore, we employ a two-step segmentation framework to preserve the objects' contours details. In the first step, the segmentation model predicts a segmentation map for the input scene image. Contour detection is performed on this scene segmentation map to obtain the bounding boxes for the objects. These bounding boxes are used to divide the scene image into multiple sub-images, each of which contains a single object. In step two, these sub-images are fed to the same trained segmentation model for predicting the individual object segmentation maps.

 \subsection{Persistent Features Generation}
 \label{featureGeneration}
 Segmentation maps are essentially binary images comprising only black and white voxels. A grayscale image, suitable for building a filtration of cubical complexes, can be generated from such binary images using various filtration functions. We select a commonly used filtration function, known as 
 %for highlighting topological features in data is 
 the height function, which computes a sufficient statistic to uniquely represent shapes in $\mathbb{R}^{2}$ and surfaces $\mathbb{R}^{3}$ in the form of the persistent homology transform \cite{turner2014persistent}. 
 
 For the cubical complexes in our case, we define the height function as stated in \cite{garin2019topological}. Consider a binary object segmentation map $\mathcal{B}: I \subseteq \mathbb{Z}^{n} \longrightarrow \left\{ 0,1 \right\}$, a grayscale image $\mathcal{H} : I \longrightarrow \mathbb{R} $ for the segmentation map $I$, and a direction $p \in \mathbb{R}^{n}$ of unit norm. 
 %is chosen as the height function. 
 The values of all the voxels of $\mathcal{H}$ are then reassigned as
\begin{equation}
\label{heightFiltration}
  \mathcal{H}(v) :=
    \begin{cases}
      \langle v,p\rangle & \text{if  $\mathcal{B}(v)=1$}\\
      H_{\infty} & \text{if  $\mathcal{B}(v)=0$}.
    \end{cases}       
\end{equation}
Here, $\langle v,p \rangle$ is the distance of voxel $v$ from the hyperplane defined by $p$ and $H_{\infty}$ is the filtration value of the voxel that is farthest away from the hyperplane. 

In step three, we obtain $d$ such grayscale images for each object segmentation map by considering $d$ directions that are evenly spaced on a unit $1$-sphere $\mathbb{S}^{1}$. We construct cubical complexes from each grayscale image according to Eq. (\ref{buildingComplex}). The sublevel sets of the $d$ complexes are then computed according to Eq. (\ref{sublevelSets}) to obtain $d$ filtrations. Persistent homology is applied to these filtrations to obtain $d$ persistence diagrams (PDs) for every object segmentation map. We only consider $0^{th}$ order homology for generating the PDs. We investigate the performance of two types of persistent features from the generated PDs. The following subsections \ref{sparsepi} and \ref{amplitude} describe their computation details.

\subsubsection{Sparse persistence image features}
\label{sparsepi}
\begin{figure}[t!]
    \centering
    %changed variable from v to p in the image and text
    \includegraphics[width=0.8\columnwidth]{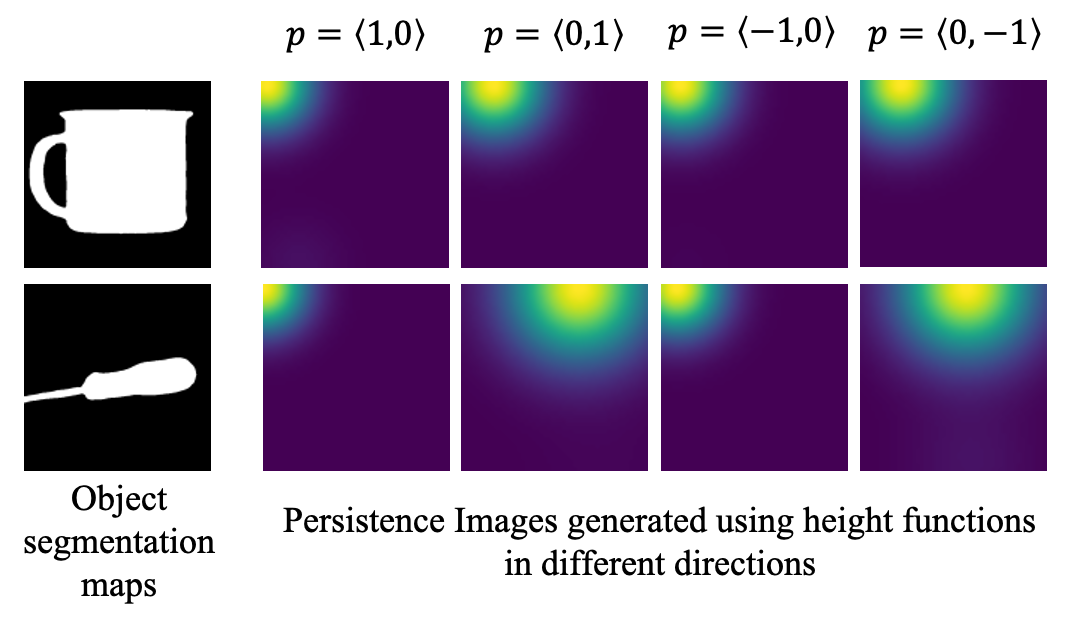}
    \caption{Persistence images (PIs) for two sample objects. The collection of PIs, computed using height functions in different directions on object segmentation maps, have enough information to distinguish between the two objects.}
    \label{PIs}
\end{figure}

Since the number of points in a PD varies from shape to shape, such a representation is not suitable for machine learning tasks. Instead, we use the persistence image (PI) representation to generate suitable features for training the recognition network. However, only a few key pixel locations of the PIs, which contain nonzero entries, sparsely encode topological information. Therefore, we adopt QR-pivoting based sparse sampling to obtain a Sparse PI, as proposed in \cite{guo2018sparse}. 
For every object segmentation mask, $d$ PIs are generated from their corresponding PDs. Fig. \ref{PIs} shows sample PIs for two different objects using height functions in multiple directions, illustrating the (collective) presence of sufficient discriminative information. In step four, for every direction $p_k, k \in \{1, \ldots, d \}$, the corresponding PIs for all the $N$ training (binary) object segmentation maps are vectorized and arranged as columns of a large matrix $X^{k}_{N}$.
%where $N$ is the number of training object segmentation maps, and $k$ indicates the $k^{th}$ direction.

The dominant PI patterns $U_{l^{k}}$ are obtained by computing the truncated singular value decomposition of $X^{k}_{N}$ as 
\begin{equation}
    X^{k}_{N} \approx U_{l^{k}}^{k}\Sigma_{l^{k}}^{k}(V_{l^{k}}^{k})^{T},
\end{equation}
where $l^{k}$ is the optimal singular value threshold \cite{gavish2014optimal} for the $p_k$th direction PIs. $U_{l^{k}}$ is then discretely sampled using the pivoted QR factorization as 
\begin{equation}
    U^{T}_{l^{k}} (\Pi^{k})^{T} = Q^{k}R^{k}.
\end{equation}
The numerically well-conditioned row permutation matrix $\Pi^{k}$ is then multiplied with $X^{k}_{N}$ to give a matrix $\bar{X}^{k}_{N}$ of sparsely sampled PIs. We finally perform row wise concatenation of all the $d$ sparsely sampled PI matrices to generate the overall set of features for recognition in step five.  
%Here
%\hl{Therefore, each column of $\bar{X}^{k}_{N}$ represents the $p_k$th direction sparse PI for the corresponding object segmentation map. For a particular object segmentation map, the sparse PIs for all the $d$ directions are then stacked to form a single feature vector for recognition.}

\subsubsection{Amplitude features}
\label{amplitude}
An alternative method of generating topologically persistent features is using the amplitude or the distance of a given PD from an empty diagram. For each of the $d$ generated PDs corresponding to an object segmentation map, we compute the bottleneck amplitude \cite{garin2019topological}, $A_{b}^{k}$, as 
\begin{equation}
    \label{bottleneckAmplitude}
    A^{k}_{b} = \frac{\sqrt{2}}{2}\sup_{j}(y^k_{j}-x^k_{j}),
\end{equation}
where $(x^k_{j},y^k_{j})$ are all the non-diagonal points in the $p_k$-th direction PD. These amplitudes are stacked to form a $d$-dimensional feature vector. Such $d$-dimensional feature vectors are generated for all the $N$ object segmentation maps to train the recognition network.

\section{Datasets}
\label{datasets}

\subsection{MPEG-7 Shape Silhouette Dataset}

\begin{figure}[t]
    \centering
    \includegraphics[width=0.75\columnwidth]{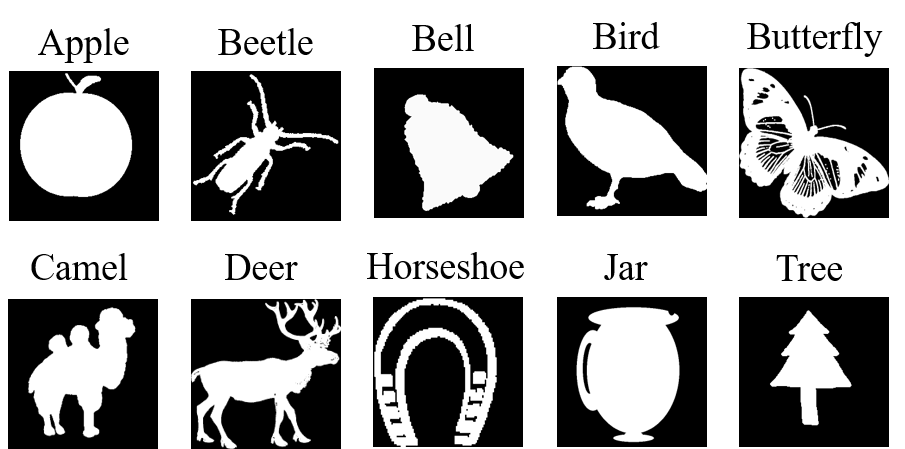}
    \caption{Sample images from the MPEG-7 Shape Silhoutte Dataset}
    \label{mpeg}
\end{figure}

We first choose the widely used MPEG-7 Shape Silhouette dataset to solely characterize the shape recognition capability of the two persistent features-based networks for (almost) ideal object segmentation maps. In particular, we use a subset of this dataset, namely, the MPEG-7 CE Shape 1 Part B dataset, which is specifically designed to evaluate the performance of 2D shape descriptors for similarity-based image retrieval \cite{latecki2000shape}. It includes the shapes of 70 different classes and 20 images for each class, for a total of 1,400 images. Fig. \ref{mpeg} shows sample images from the dataset. 

\subsection{RGB-D Scenes Dataset v1}

The shapes in the MPEG-7 are detailed and fairly distinguishable from each other. However, common objects in indoor environments are often less detailed and, therefore, more challenging for topological methods. Most deep learning models work exceptionally well in recognizing such everyday objects in their training environments. However, they face challenges when used in new (previously unseen) environments without any retraining, even if the objects remain the same. Therefore, we choose the widely used benchmark, the RGB-D Scene Dataset v1\cite{lai2011large}, to evaluate the performance of our proposed methods. The dataset consists of eight scene setups with objects that belong to the RGB-D Object Dataset\cite{lai2011large}. The scenes are shot in five different environments: \textit{desk}, \textit{kitchen\_small}, \textit{meeting\_small}, \textit{table}, and \textit{table\_small}. In particular, we select the \textit{table\_small} and \textit{desk} environments for our evaluation. Among all the objects present in the scenes, we consider six object classes for recognition corresponding to the six objects types that appear in both the environments. For our current analysis, we only consider RGB images where the objects are not occluded. 

\vspace{1mm}
\subsection{UW Indoor Scenes Dataset}

While the RGB-D Scenes Dataset consists of scenes from multiple environments, all of them are tabletop environments with limited variation in terms of lighting conditions and object types. On the other hand, the dataset in \cite{rennie2016dataset} provides images with several objects but only in a single environment. Although there are quite a few other indoor scene datasets in the literature, none of them includes a large enough set of object types, poses, and arrangements with varying backgrounds and lighting conditions. Therefore, we introduce a new RGB dataset, which we call the UW Indoor Scenes (UW-IS) dataset, for evaluating object recognition performance in multiple, typical indoor environments. For this purpose, we select a fixed set of fourteen different objects from the benchmark Yale-CMU-Berkeley (YCB) object and model set \cite{calli2015benchmarking}. 

%scenes with different training and test 
%Additionally, scene-level segmentation maps are not available. 
%my original reason for not using the dataset from Dieter Fox's lab [28] was because it did not have segmentation maps. It also has a lot of clutter (and we weren't dealing with occlusions at this stage. Absence of segmentation maps was also another reason; but then we did labeling anyway (for our dataset) so its probably not a good reason. I didn't want to bring attention to clutter and occlusion in the paper anywhere else apart from future work. Therefore, I dug deeper to come up with a better 'paper-worthy' reason. The most recent version of the dataset consists of 22 scenes (22 videos essentially) but they consist of a maximum of 5 categories of objects apart from furniture objects like sofas, chair (they sometimes count these as objects sometimes depending on application).

\begin{figure}[t]
\begin{subfigure}{\columnwidth}
  \centering
  \includegraphics[width=0.8\textwidth]{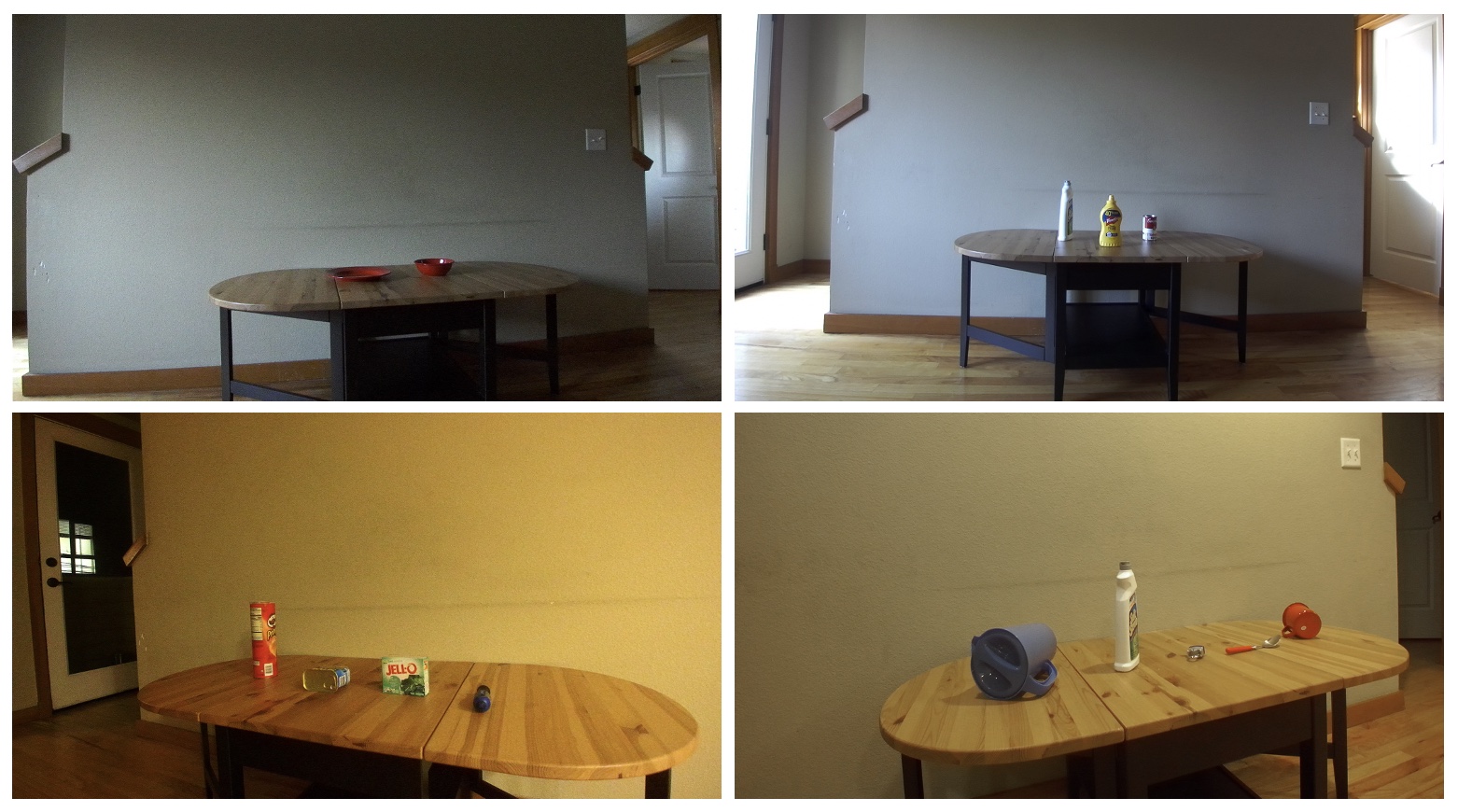} 
  \caption{Living room environment}
  \label{livingroom}
\end{subfigure}

\begin{subfigure}{\columnwidth}
  \centering
  \includegraphics[width=0.8\textwidth]{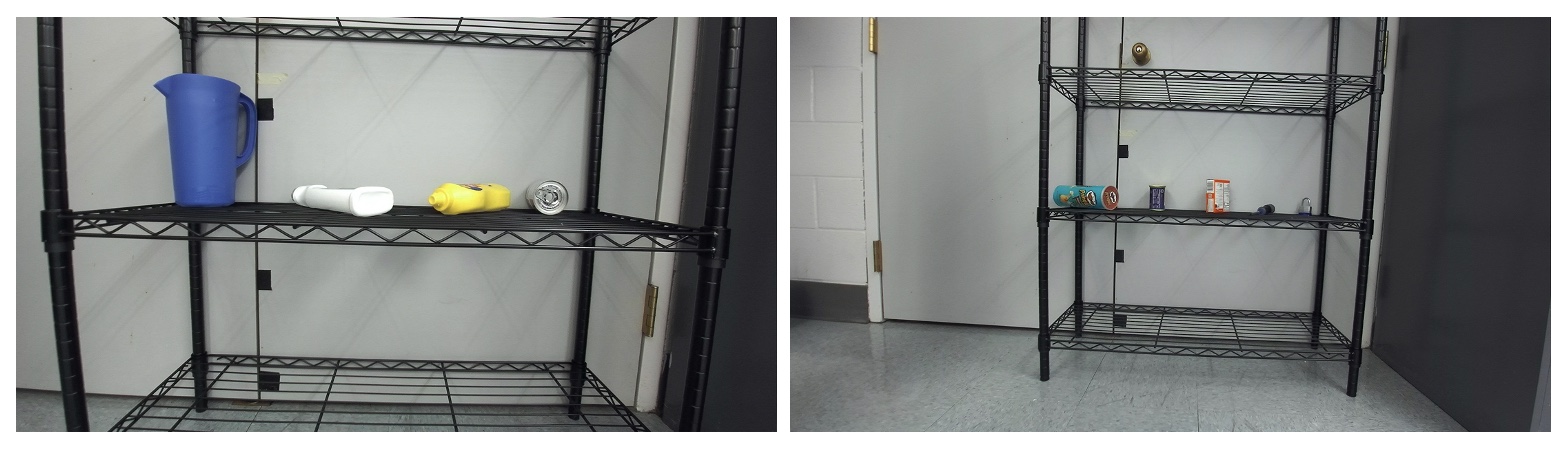}  
  \caption{Mock warehouse environment}
  \label{warehouse}
\end{subfigure}

\begin{subfigure}{\columnwidth}
  \centering
  \includegraphics[width=0.8\textwidth]{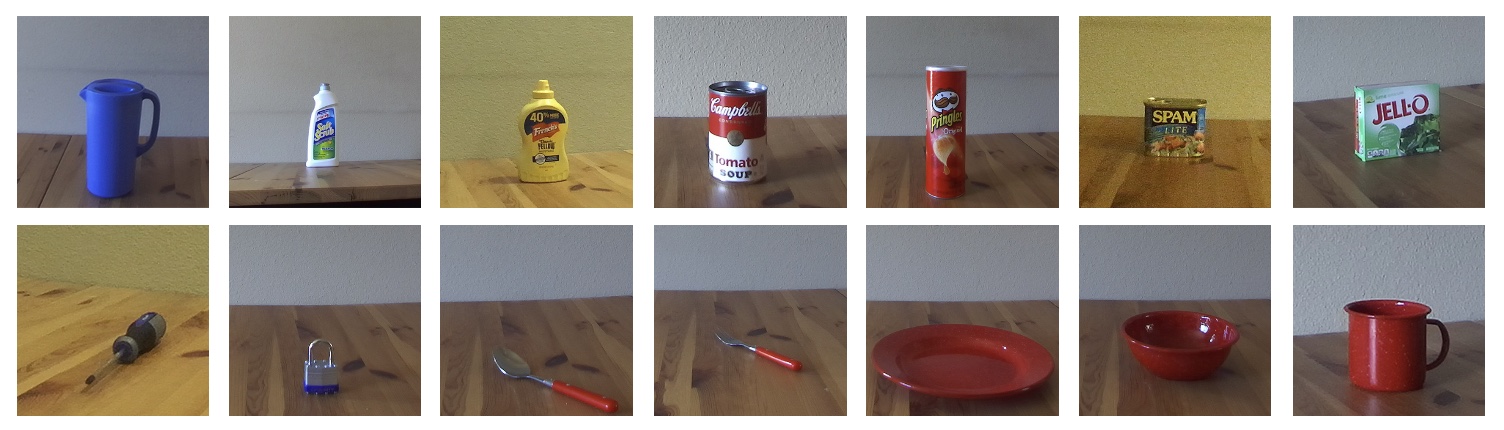}  
  \caption{Objects}
  \label{objects}
\end{subfigure}

\caption{Representative images from the UW Indoor Scenes Dataset}
\label{dataset}
\end{figure}

The UW-IS dataset consists of indoor scenes taken in two completely different environments. The first environment is a living room scene where the objects are placed on a tabletop. The second environment is a mock warehouse setup where the objects are placed on a shelf. For the living room environment, we have a total of 347 scene images. The images are taken in four different illumination settings from three different camera perspectives and varying distances up to two meters. Sixteen out of the 347 scene images are with two different objects, 135 images are with three different objects, 156 images are with four different objects, and 40 images are with five different objects. For the mock warehouse environment, we have a total of 200 scene images taken from distances up to 1.6 meters. Sixty out of 200 images are images with three different objects, 68 images with four different objects, and 72 images are with five different objects. Fig. \ref{livingroom} shows some sample living room scene images, and Fig. \ref{warehouse} shows sample images from the mock warehouse environment. Fig. \ref{objects} shows all the fourteen objects used in our dataset. The dataset is publicly available at \href{https://data.mendeley.com/datasets/dxzf29ttyh/}{\url{https://data.mendeley.com/datasets/dxzf29ttyh/}}.

%\vspace{-1mm}
\section{Experiments}
\label{experiments}

\subsection{Implementation Details}
We perform five-fold training and testing on all images of the MPEG-7 Shape Dataset. For evaluation on the RGB-D Scenes dataset, we use the \textit{table\_small} environment for both training and testing, and the \textit{desk} environment only for testing. For the UW-IS dataset, we use the living room images for training and testing, and the mock warehouse images only for testing. All the training and testing is performed using GeForce GTX 1080 and 1080 Ti GPUs on workstations running Windows 10 and Ubuntu 18.04 LTS, respectively. The code for the proposed methods is available at \href{https://github.com/smartslab/objectRecognitionTopologicalFeature}{\url{https://github.com/smartslab/objectRecognitionTopologicalFeatures}}.

For the MPEG-7 Shape Dataset, we divide the 1,400 images into five sets of 280 images each, with four images of each class included in every set. We perform five-fold training and testing using these sets, such that each set is used once as a test set while the remaining four sets are used for training and validation. We use the giotto-tda \cite{tauzin2020giottotda} library to generate the PDs and the Persim package in Scikit-TDA Toolbox to generate the PIs. The PDs are generated using height functions along 8 evenly spaced directions on $S^{1}$. We choose a grid size equal to 50 $\times$ 50, a spread of 10, and a linear weighting function for generating the PIs. We use three-layered and five-layered fully connected networks for recognition using amplitude features and sparse PI features, respectively. We use the ReLU activation (last layer uses softmax activation), Adam optimizer and categorical cross-entropy loss function for training. The learning rate is set to 0.01 for the first 500 epochs. It is decreased by a factor of 10 after every 100 epochs for the next 200 epochs, and by a factor of 100 for the last 100 epochs.

For the RGB-D Scenes dataset and the UW-IS dataset, we use the DeepLabv3+ segmentation model with Xception-65 backbone following the implementation in \cite{tensorflowdeeplab} to obtain object segmentation maps. The network is initialized using a model pre-trained on the ImageNet and PASCAL VOC2012 datasets available from \cite{tensorflowdeeplab}. We use 147 images from the \textit{table\_small} environment to train the network for the RGB-D Scenes dataset and 200 living room images to train the network for the UW-IS dataset. Horizontally flipped counterparts of the images are also used for training the models. We generate the corresponding foreground annotations using LabelMe \cite{wkentaro_2019}. We train the network for 20,000 steps using the categorical cross-entropy loss with 10\% hard example mining after 2,500 steps in the case of the RGB-D Scenes dataset. For the UW-IS dataset, we use 1\% hard example mining\footnote{Unlike the UW-IS dataset, the smaller training set for the RGB-D Scenes dataset leads to poor quality object segmentation maps that are unsuitable for extracting topological features. Therefore, we use a separate DeepLabv3+ model for Step 2, which is fine-tuned on images of individual objects obtained from the original scene images}.

We then perform five-fold training and testing of the proposed methods on both datasets by diving the scene images from the training environment into five folds. The resulting five models are also tested separately on all the images of the test environment. First, the trained DeepLabv3+ segmentation models are used to obtain object segmentation maps from all the images. To ensure the quality of segmentation maps remain consistent across training and test environments, we generate object segmentation maps for the test environments using segmentation models fine-tuned in these environments. We then appropriately pad all the segmentation maps with zeros and consistently resize them to obtain 125 $\times$ 125 binary images. We augment the training data by rotating every object segmentation map by $90^{o}$, $180^{o}$, and $270^{o}$. The PDs and PIs are generated in the same manner as for the MPEG-7 dataset, except that the spread value is chosen to be 20. We also use the same fully connected network architectures and hyperparameters as for the MPEG-7 dataset.

\textbf{Comparison with deep learning-based object recognition methods:} 
To solely characterize the recognition performance of the topologically persistent features, we compare their performance with features extracted using two other widely used object recognition methods, namely, ResNetV2-56\cite{he2016identity} and EfficientNet-B4\cite{tan2019efficientnet}, on the RGB-D Scenes benchmark. We use the same sets of cropped objects obtained using DeepLabv3+ in Step 1 of our proposed method for five-fold training and testing of both the methods. The networks for both the methods are trained using the implementations available in Keras \cite{keras}.

\textbf{Comparison with end-to-end object detection methods:} We also compare the overall performance of both the persistent features-based methods against end-to-end object detection methods on both the RGB-D Scenes dataset and the UW-IS dataset. Particularly, we compare performance against Faster R-CNN\cite{ren2015faster}, a widely used object detection method, and its state-of-the-art variant for cross-domain object detection, Domain Adaptive Faster R-CNN \cite{chen2018domain}. For Faster R-CNN (referred to as FR-CNN), we use a pre-trained model with InceptionResNet-V2 feature extractor and hyperparameters available with the TensorFlow Object Detection API\cite{huang2017speed}. For Domain Adaptive Faster R-CNN, we use a modified implementation from \cite{krumo}, where the VGG backbone is replaced with ResNet-50, and the RoI-pooling layer is replaced with the more popular RoIAlign. The modified implementation has been shown to outperform other state-of-the-art methods for cross-domain object detection\cite{krumo}. We call this improved method DA-FR-CNN\textbf{*}. Similar to the proposed methods, we perform five-fold training and testing of these methods on both datasets. In the case of DA-FR-CNN\textbf{*} for the UW-IS dataset, all the training set images with artificial lighting (e.g., bottom row in Fig. \ref{livingroom}) are used as the source domain, while all the images with natural lighting (e.g., top row in Fig. \ref{livingroom}) are used as the target domain. Since such a split is not possible in the RGB-D Scenes dataset, we randomly divide the images into the source and target domains. The ground truth bounding boxes for both the datasets are generated using LabelImg\cite{tzutalin}.

\subsection{Results}

We first examine the recognition performance of both amplitude features and sparse PI features on the MPEG-7 dataset. We use the weighted F1 score, weighted precision, weighted recall, and accuracy for evaluating performance. Table \ref{mpegtable} shows the test-time performance of the trained recognition networks. We observe that sparse PI-based recognition performance is quite impressive (more than 0.85) with respect to all the four metrics. It is also consistently better than amplitude-based recognition, indicating the usefulness of sparse sampling in selecting the key features. It is worth noting that 100\% recognition performance using 2D shape knowledge is not possible for this dataset, since some of the classes contain shapes that are significantly different from the others in the same class but are similar to certain shapes in other classes \cite{latecki2000shape}.
%with respect to all the four reported metrics. In fact, sparse PI-based recognition achieves an accuracy of $0.87\pm0.01$. 

\begin{table}[h]
%updated with std errors, updated metrics
\centering
\caption{Performance comparison of amplitude and sparse PI features (best in bold) on the MPEG Shape Silhouette dataset. (w) indicates weighted metric.}
\label{mpegtable}
\resizebox{0.6\columnwidth}{!}{%
\begin{tabular}{@{}ccc@{}}
\toprule
               & Amplitude & Sparse PI \\ \midrule
F1 score (w)   & 0.75$\pm$0.01 & \textbf{0.87$\pm$0.01} \\
Precision (w) & 0.77$\pm$0.01 & \textbf{0.89$\pm$0.02} \\
Recall (w)     & 0.76$\pm$0.01 & \textbf{0.87$\pm$0.01} \\
Accuracy & 0.76$\pm$0.01 & \textbf{0.87$\pm$0.01} \\ \bottomrule
\end{tabular}%
}
\end{table}

\begin{table*}[h]
\centering
\caption{Performance comparison of the proposed persistent features-based methods with ResNetV2-56, EfficientNet-B4, FR-CNN, and DA-FR-CNN\textbf{*} (best in bold) on the \textit{desk} images from the RGB-D Scenes dataset.}
\label{uwrgbdTable}
\resizebox{\textwidth}{!}{%
\begin{tabular}{@{}cccc|cc|cc@{}}
\toprule
Metric                    & Class & Amplitude & Sparse PI & ResNetV2-56 & EfficientNet-B4 & FR-CNN & DA-FR-CNN* \\ \midrule
\multirow{6}{*}{F1 score} & Bowl  & 0.71$\pm$0.01 & 0.69$\pm$0.01 & 0.95$\pm$0.01  & 0.90$\pm$0.03       & 0.83$\pm$0.02   & 0.88$\pm$0.01 \\
              & Cap        & 0.36$\pm$0.02 & 0.63$\pm$0.03 & 0.46$\pm$0.06 & 0.58$\pm$0.04 & 0.46$\pm$0.03 & 0.99$\pm$0.01 \\
              & Cereal box & 0.80$\pm$0.01 & 0.88$\pm$0.01 & 0.76$\pm$0.03 & 0.83$\pm$0.03 & 0.65$\pm$0.01 & 0.79$\pm$0.01 \\
              & Cup        & 0.24$\pm$0.02 & 0.50$\pm$0.03 & 0.11$\pm$0.08 & 0.06$\pm$0.04 & 0.87$\pm$0.02 & 0.26$\pm$0.12 \\
              & Soda can   & 0.76$\pm$0.01 & 0.66$\pm$0.01 & 0.30$\pm$0.05 & 0.29$\pm$0.04 & 0.62$\pm$0.01 & 0.51$\pm$0.03 \\
              & Stapler    & 0.70$\pm$0.01 & 0.71$\pm$0.01 & 0.88$\pm$0.03 & 0.84$\pm$0.05 & 0.69$\pm$0.03 & 0.78$\pm$0.01 \\
\midrule
F1 score (w)  & -          & 0.70$\pm$0.00 & 0.71$\pm$0.01 & 0.65$\pm$0.01 & 0.65$\pm$0.01 & 0.70$\pm$0.01 & \textbf{0.73$\pm$0.01} \\
Precision (w) & -          & 0.75$\pm$0.01 & 0.74$\pm$0.00 & 0.79$\pm$0.01 & 0.77$\pm$0.01 & 0.82$\pm$0.01 & \textbf{0.87$\pm$0.01} \\
Recall (w)    & -          & 0.68$\pm$0.00 & \textbf{0.71$\pm$0.01} & 0.68$\pm$0.01 & 0.69$\pm$0.01 & 0.66$\pm$0.01 & 0.69$\pm$0.01 \\
Accuracy      & -          & 0.68$\pm$0.00 & \textbf{0.71$\pm$0.01} & 0.68$\pm$0.01 & 0.69$\pm$0.01 & 0.66$\pm$0.01 & 0.69$\pm$0.01 \\ \bottomrule
\end{tabular}%
}
\end{table*}

We then examine the performance\footnote{We do not account for the false negatives and false positives resulting from errors of the segmentation model to ensure a fair judgement of our methods' effectiveness. Equivalently, for object detection methods, false positives arising from incorrect region proposals are also not considered.} of both the persistent features-based methods along with other object recognition and detection methods on the RGB-D Scenes benchmark. Table \ref{uwrgbdTable} reports the performance of all the methods (trained in the \textit{table\_small} environment) on the unseen \textit{desk} environment. We observe that sparse PI-based recognition, which achieves an overall accuracy of $0.71\pm0.01$, has the highest performance among all the methods in terms of recall and accuracy. Particularly, for the same set of object images obtained using the DeepLabv3+ model, recognition using sparse PI features computed from object segmentation maps is better than recognition using features learned by ResNetV2-56 and EfficientNet-B4 from RGB inputs with respect to accuracy, recall, and F1-score. Additionally, amplitude features-based recognition is comparable to both ResNetV2-56 and EfficientNet-B4. Additionally, the difference between the recognition performance using sparse PI features and amplitude features is, however, much lower than that for the MPEG-7 dataset. We also note that the sparse PI features-based method outperforms Faster R-CNN (referred to as FR-CNN) substantially and DA-FR-CNN\textbf{*} by a small margin in terms of accuracy and recall.

We then assess the performance of both the persistent features-based methods, FR-CNN, and DA-FR-CNN\textbf{*} on the living room scene images from the UW-IS dataset reported in Table \ref{livingRoomTable}. We observe that both the persistent features-based methods, which only use the information in object segmentation maps, perform reasonably well. Sparse PI-based recognition achieves an overall accuracy of $0.71\pm0.01$ and is marginally better than amplitude-based recognition, whose accuracy is $0.69 \pm 0.01$. Moreover, both FR-CNN and DA-FR-CNN\textbf{*}, which are trained on this environment and use RGB images as inputs, outperform these methods with respect to all the metrics including class-wise F1 scores.

\begin{table}[h]
\centering
\caption{Performance comparison of the proposed persistent features-based methods with FR-CNN and DA-FR-CNN\textbf{*} (best in bold) on the living room images from the UW-IS dataset.}

%say F RCNN instead of Faster RCNN and spell it out a couple of times
\label{livingRoomTable}
\resizebox{\columnwidth}{!}{%
\begin{tabular}{@{}cccccc@{}}
\toprule
Metric                     & Class           & Amplitude & Sparse PI & FR-CNN & DA-FR-CNN\textbf{*} \\ \midrule
\multirow{14}{*}{F1 score} & Spoon           & 0.62$\pm$0.04 & 0.57$\pm$0.01 & 0.84$\pm$0.03    & 0.68$\pm$0.02 \\
                           & Fork            & 0.15$\pm$0.05 & 0.16$\pm$0.07 & 0.81$\pm$0.02   & 0.32$\pm$0.06 \\
                           & Plate           & 0.83$\pm$0.02 & 0.89$\pm$0.01 & 0.98$\pm$0.01   & 0.95$\pm$0.02 \\
                           & Bowl            & 0.95$\pm$0.01 & 0.94$\pm$0.02 & 0.98$\pm$0.01   & 0.97$\pm$0.01 \\
                           & Cup             & 0.67$\pm$0.02 & 0.78$\pm$0.04 & 0.91$\pm$0.01   & 1.00$\pm$0.00 \\
                           & Pitcher base    & 0.81$\pm$0.03 & 0.87$\pm$0.02 & 0.92$\pm$0.02   & 0.98$\pm$0.01\\
                           & Bleach cleanser & 0.73$\pm$0.02 & 0.72$\pm$0.02 & 0.87$\pm$0.03   & 0.97$\pm$0.01 \\
                           & Mustard bottle  & 0.63$\pm$0.02 & 0.68$\pm$0.03 & 0.84$\pm$0.03   & 0.98$\pm$0.01 \\
                           & Soup can & 0.68$\pm$0.04 & 0.69$\pm$0.02 & 0.84$\pm$0.04   & 0.86$\pm$0.02 \\
                           & Chips can       & 0.72$\pm$0.02 & 0.75$\pm$0.02 & 0.92$\pm$0.03   & 0.90$\pm$0.03 \\
                           & Meat can & 0.55$\pm$0.05 & 0.56$\pm$0.04 & 0.82$\pm$0.03   & 0.88$\pm$0.01 \\
                           & Gelatin box     & 0.55$\pm$0.02 & 0.64$\pm$0.04 & 0.91$\pm$0.02   & 0.91$\pm$0.01 \\
                           & Screwdriver     & 0.58$\pm$0.04 & 0.72$\pm$0.04 & 0.91$\pm$0.02   & 0.94$\pm$0.02 \\
                           & Padlock         & 0.74$\pm$0.03 & 0.72$\pm$0.03 & 0.97$\pm$0.02   & 0.93$\pm$0.03 \\
\midrule
F1 score (w)               & -               & 0.68$\pm$0.01 & 0.71$\pm$0.01 & \textbf{0.89$\pm$0.01}  & 0.88$\pm$0.01 \\
Precision (w)              & -               & 0.69$\pm$0.01 & 0.71$\pm$0.02 & \textbf{0.91$\pm$0.01}   & 0.90$\pm$0.01 \\
Recall (w)                 & -               & 0.69$\pm$0.01 & 0.71$\pm$0.01 & \textbf{0.88$\pm$0.01}   & \textbf{0.88$\pm$0.00}\\
Accuracy                   & -               & 0.69$\pm$0.01 & 0.71$\pm$0.01 & \textbf{0.88$\pm$0.01}   & \textbf{0.88$\pm$0.00} \\ \bottomrule
\end{tabular}%
}
\end{table}

We believe that the quality of the object segmentation maps has substantial impact on the performance of the persistent features-based methods. Notably, performance is considerably worse for the fork as compared to the other objects. To assess this impact, we compare the recognition performance of both these methods with that of a human given an identical set of segmentation maps. 
%We obtain the object recognition performance of a human on the generated object segmentation maps and compare our methods' performance with human performance.
Table \ref{humanPerformance} summarizes the comparison results. For the persistent features based-methods, we only report the accuracy for those images where the human recognizes the object correctly. We observe that a human achieves an accuracy of $0.84\pm0.01$, which is lower that FR-CNN performance, and finds it difficult to recognize the objects based on the generated segmentation maps, especially for the spoon and fork classes. We refer the reader to Section \ref{discussion} for further discussion on segmentation map quality.

\begin{table}[ht]
%updated with std errors
\centering
\caption{Comparison of recognition accuracy %performance
of the proposed persistent features-based methods with a human given the object segmentation maps for living room scene images from the UW-IS dataset}
\label{humanPerformance}
\resizebox{0.85\columnwidth}{!}{%
\begin{tabular}{@{}cccc@{}}
\toprule
Class           & Human performance & Amplitude & Sparse PI \\ \midrule
Spoon           & 0.37$\pm$0.07         & 0.44$\pm$0.05 & 0.47$\pm$0.09 \\
Fork            & 0.40$\pm$0.11         & 0.16$\pm$0.08 & 0.12$\pm$0.07 \\
Plate           & 0.96$\pm$0.02         & 0.90$\pm$0.03 & 0.90$\pm$0.04 \\
Bowl            & 0.97$\pm$0.01         & 0.96$\pm$0.01 & 0.97$\pm$0.01 \\
Cup             & 0.92$\pm$0.03         & 0.65$\pm$0.05 & 0.81$\pm$0.03 \\
Pitcher base    & 0.97$\pm$0.01         & 0.87$\pm$0.04 & 0.92$\pm$0.02 \\
Bleach cleanser & 0.91$\pm$0.01         & 0.86$\pm$0.02 & 0.77$\pm$0.02 \\
Mustard bottle  & 0.91$\pm$0.03         & 0.63$\pm$0.04 & 0.73$\pm$0.04 \\
Soup can        & 0.76$\pm$0.05         & 0.83$\pm$0.06 & 0.81$\pm$0.06 \\
Chips can       & 0.89$\pm$0.04         & 0.81$\pm$0.04 & 0.74$\pm$0.07 \\
Meat can        & 0.86$\pm$0.03         & 0.55$\pm$0.07 & 0.61$\pm$0.03 \\
Gelatin box     & 0.88$\pm$0.04         & 0.59$\pm$0.03 & 0.56$\pm$0.07 \\
Screwdriver     & 0.83$\pm$0.03         & 0.63$\pm$0.06 & 0.74$\pm$0.06 \\
Padlock         & 0.85$\pm$0.04         & 0.81$\pm$0.05 & 0.82$\pm$0.03 \\
\midrule
Accuracy          & 0.84$\pm$0.01         & 0.74$\pm$0.01 & 0.76$\pm$0.01 \\ \bottomrule
\end{tabular}%
}
\end{table}

Despite this challenge, we expect recognition performance to be relatively unaffected, when the environments vary considerably but the objects are identical, provided the quality of the segmentation maps remains consistent. Accordingly, we test the performance of the persistent features-based methods on all the warehouse scene images of the UW-IS dataset without retraining the recognition networks. We also test the performance of FR-CNN and DA-FR-CNN\textbf{*} on all the warehouse images without any retraining. Table \ref{warehouseTable} summarizes the performances of all the four methods on the mock warehouse test environment. We observe that the performance of sparse PI-based recognition is unchanged from the living room scenario, and is substantially better than that of amplitude-based recognition, whose accuracy reduces by 4\%. However, the performance of FR-CNN degrades a lot without fine-tuning (accuracy drops by 25\%). The performance of DA-FR-CNN\textbf{*} also degrades considerably (accuracy drops by 18\%). Similar to the RGB-D Scenes benchmark case, the sparse PI features-based method outperforms FR-CNN substantially and DA-FR-CNN\textbf{*} by a small margin with respect to accuracy and recall.

\begin{table}
\centering
\caption{Performance comparisons of the proposed persistent features-based methods with FR-CNN and DA-FR-CNN\textbf{*} (best in bold) on mock warehouse scene images from the UW-IS dataset.
%The bold numbers indicate cases where one method performs significantly better than the others (p-value \textless 0.05)
}
\label{warehouseTable}
\resizebox{\columnwidth}{!}{%
\begin{tabular}{@{}cccccc@{}}

\toprule
Metric                     & Class           & Amplitude & Sparse PI & FR-CNN & DA-FR-CNN\textbf{*} \\ \midrule
\multirow{14}{*}{F1 score} & Spoon           & 0.44$\pm$0.04 & 0.52$\pm$0.02 & 0.39$\pm$0.03   & 0.32$\pm$0.05 \\
                           & Fork            & 0.28$\pm$0.04 & 0.20$\pm$0.04 & 0.24$\pm$0.03   & 0.00$\pm$0.00 \\
                           & Plate           & 0.16$\pm$0.07 & 0.04$\pm$0.02 & 0.27$\pm$0.16   & 0.81$\pm$0.01\\
                           & Bowl            & 0.94$\pm$0.00 & 0.91$\pm$0.01 & 0.97$\pm$0.01   & 0.97$\pm$0.00 \\
                           & Cup             & 0.51$\pm$0.02 & 0.90$\pm$0.01 & 0.94$\pm$0.01   & 0.92$\pm$0.00 \\
                           & Pitcher base    & 0.86$\pm$0.01 & 0.94$\pm$0.00 & 0.93$\pm$0.03   & 1.00$\pm$0.00 \\
                           & Bleach cleanser & 0.79$\pm$0.02 & 0.78$\pm$0.01 & 0.84$\pm$0.04   & 0.96$\pm$0.01 \\
                           & Mustard bottle  & 0.76$\pm$0.02 & 0.74$\pm$0.01 & 0.36$\pm$0.05   & 1.00$\pm$0.00 \\
                           & Soup can        & 0.73$\pm$0.02 & 0.66$\pm$0.02 & 0.85$\pm$0.02   & 0.81$\pm$0.01 \\
                           & Chips can       & 0.76$\pm$0.01 & 0.74$\pm$0.01 & 0.71$\pm$0.03   & 0.87$\pm$0.02 \\
                           & Meat can        & 0.58$\pm$0.09 & 0.71$\pm$0.01 & 0.75$\pm$0.05   & 0.74$\pm$0.04\\
                           & Gelatin box     & 0.23$\pm$0.04 & 0.25$\pm$0.02 & 0.70$\pm$0.03   & 0.70$\pm$0.03 \\
                           & Screwdriver     & 0.78$\pm$0.02 & 0.72$\pm$0.01 & 0.42$\pm$0.04   & 0.76$\pm$0.03\\
                           & Padlock         & 0.61$\pm$0.03 & 0.84$\pm$0.01 & 0.89$\pm$0.03  & 0.60$\pm$0.04 \\
\midrule
F1 score (w)               & -               & 0.65$\pm$0.01 & 0.70$\pm$0.00 & 0.65$\pm$0.01   & \textbf{0.75$\pm$0.01} \\
Precision (w)              & -               & 0.67$\pm$0.01 & 0.70$\pm$0.01 & 0.77$\pm$0.02   & \textbf{0.87$\pm$0.01} \\
Recall (w)                 & -               & 0.66$\pm$0.01 & \textbf{0.71$\pm$0.01} & 0.63$\pm$0.01   & 0.70$\pm$0.01 \\
Accuracy                   & -               & 0.66$\pm$0.01 & \textbf{0.71$\pm$0.01} & 0.63$\pm$0.01   & 0.70$\pm$0.01 \\ \bottomrule
\end{tabular}%
}
\end{table}

\subsection{Robot Implementation}

\begin{figure}[ht]
    \centering
    \includegraphics[width=0.7\columnwidth]{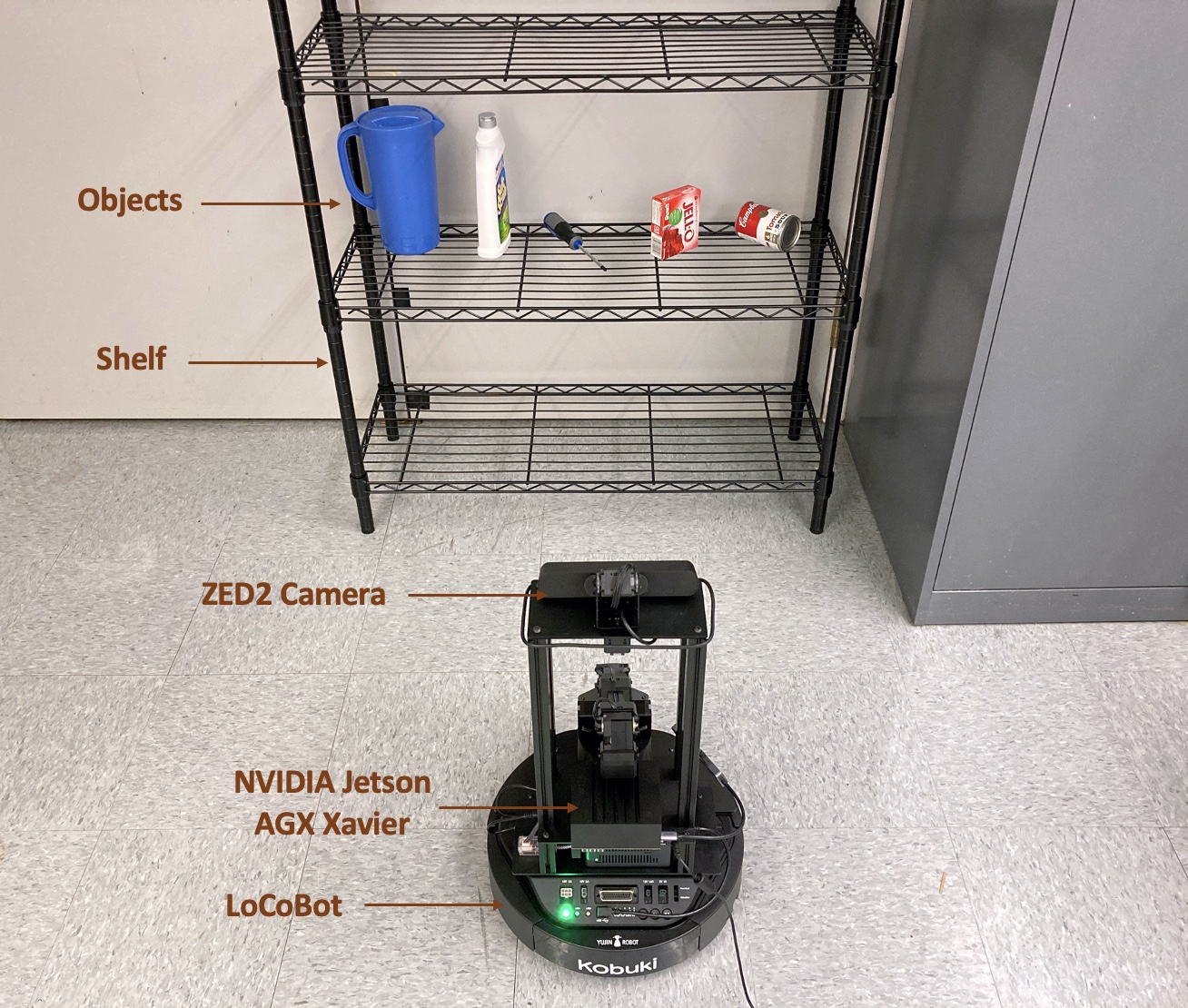}
    \caption{Screenshot of the LoCoBot operating in a mock warehouse setup}
    \label{robot}
\end{figure}

We also implement our proposed framework on a LoCoBot platform built on a Yujin Robot Kobuki Base (YMR-K01-W1) and powered by an Intel NUC NUC7i5BNH Mini PC. We mount a ZED2 camera with stereo vision on top of the LoCoBot and control the robot using the PyRobot interface \cite{pyrobot2019}. The camera images are fed to the trained segmentation model and recognition networks, which are run on an on-board NVIDIA Jetson AGX Xavier processor, equipped with a 512-core Volta GPU with Tensor Cores and 8-core ARM v8.2 64-bit CPU. We use TensorRT \cite{nvidia} for optimizing the trained segmentation model. Fig. \ref{robot} shows a screenshot of the platform. The sparse PI features-based recognition runs at a speed of $1.1s$ per frame on this platform. A video demonstration of object recognition on this platform is included in the Supplementary Materials. 

%\vspace{-7mm}
\section{Discussion}
\label{discussion}

\begin{figure}[t]
\centering
\begin{subfigure}{0.23\columnwidth}
\includegraphics[width=\linewidth]{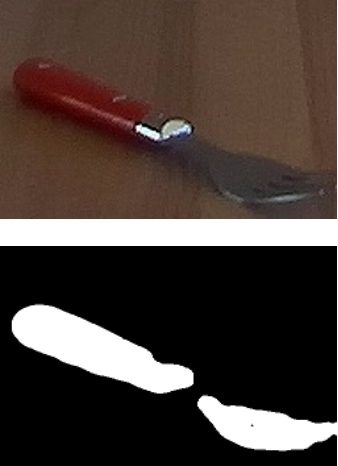}
\caption{}\label{fork}
\end{subfigure}
\begin{subfigure}{0.15\columnwidth}
\includegraphics[width=\linewidth]{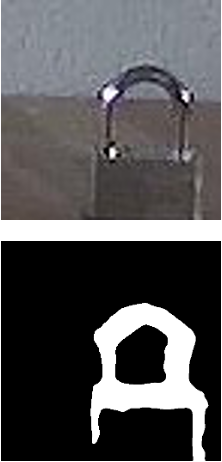}
\caption{}\label{lock}
\end{subfigure}
\begin{subfigure}{0.3\columnwidth}
\includegraphics[width=\linewidth]{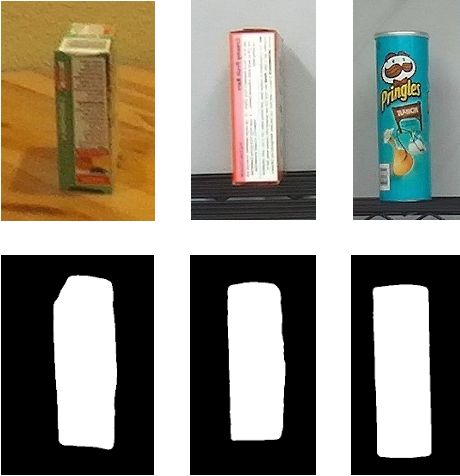}
\caption{}\label{gelatinbox}
\end{subfigure}
\smallskip
\begin{subfigure}{0.7\columnwidth}
\includegraphics[width=\linewidth]{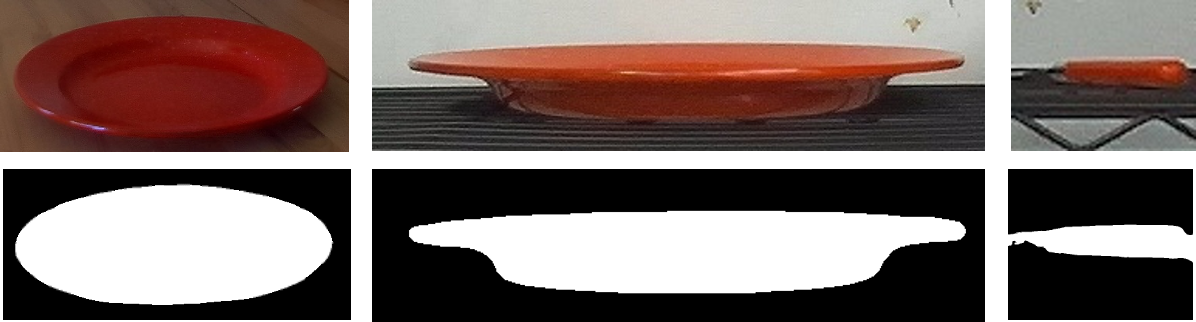}
\caption{}\label{plate}
\end{subfigure}

\caption{Sample failure cases. Fig. \ref{fork} shows a case where a poorly segmented fork is confused with a spoon. Fig. \ref{lock} shows a case where, unlike the proposed topological methods, a human succeeds by being able to complete the shape. Fig. \ref{gelatinbox} and \ref{plate} illustrate that camera pose changes cause some of the objects to appear similar to other objects.}
\label{failure}
\end{figure}

We note that segmentation map quality has a considerable impact on the performance of the persistent features-based methods. For example, the performance is particularly bad for the fork and spoon classes, as observed from Tables \ref{livingRoomTable} and \ref{warehouseTable}. The inherent similarity between forks and spoons, along with low segmentation quality, often makes it hard to distinguish between them, as shown in Fig. \ref{fork}. Table \ref{humanPerformance} shows that even humans have difficulty distinguishing between them from the generated segmentation maps. Both Faster R-CNN (referred to as FR-CNN) and DA-FR-CNN\textbf{*} also have difficulties with these classes, especially in the unseen warehouse environment. On a related note, incomplete segmentation maps also affect performance of the proposed methods. For example, the shape in the incomplete segmentation map of a padlock shown in Fig. \ref{lock} is quite different from that of a padlock, making it difficult for a topology-based method to recognize the object. Moreover, we observe from Table \ref{humanPerformance} that our methods only achieve 76\% and 74\% of human performance using sparse PI and amplitude features, respectively. We believe this difference is primarily due to humans' additional cognitive capability to complete (partially visible) shapes.

Additionally, we observe relatively small variations in the class-wise performance of the proposed methods between the living room and mock warehouse environments except for the plate and gelatin box classes. This observation can be attributed to changes in camera locations and viewing angles that lead to unseen object poses and naturally occurring variations in object placements. Fig. \ref{gelatinbox} illustrates this problem for the gelatin box. The leftmost image is from the living room. The middle image shows the same object pose captured from a different camera viewing angle in the warehouse. The 2D shape in the middle image becomes very similar to that of the chips can in the rightmost image. Similarly, Fig. \ref{plate} shows how such a change results in a completely different 2D shape for the plate, which is similar to a half-visible spoon. The performance of FR-CNN is also affected by such variations in camera pose, as observed for the plate class in Table \ref{warehouseTable}. On the other hand, the performance of DA-FR-CNN\textbf{*} drops due to changes in object appearance. For example, the cups in the \textit{desk} and \textit{table\_small} environments look considerably different, leading to poor recognition, as reported in Table \ref{uwrgbdTable}.

% \begin{figure}[t]
% \centering
% \begin{subfigure}{0.247\columnwidth}
% \includegraphics[width=\linewidth]{figures/disc_subfigb.png}
% \caption{}\label{fork}
% \end{subfigure}
% \begin{subfigure}{0.161\columnwidth}
% \includegraphics[width=\linewidth]{figures/disc_subfiga.png}
% \caption{}\label{lock}
% \end{subfigure}
% \begin{subfigure}{0.322\columnwidth}
% \includegraphics[width=\linewidth]{figures/disc_subfigc.png}
% \caption{}\label{gelatinbox}
% \end{subfigure}
% \smallskip
% \begin{subfigure}{0.75\columnwidth}
% \includegraphics[width=\linewidth]{figures/disc_subfigd.png}
% \caption{}\label{plate}
% \end{subfigure}

% \caption{Sample failure cases. Fig. \ref{fork} shows a case where a poorly segmented fork is confused with a spoon. Fig. \ref{lock} shows a case where, unlike the proposed topological methods, a human succeeds by being able to complete the shape. Fig. \ref{gelatinbox} and \ref{plate} illustrate that camera pose changes cause some of the objects to appear similar to other objects.}
% \label{failure}
% \end{figure}

\addtolength{\textheight}{-0.2cm}
%\vspace{-1mm} 
\section{CONCLUSIONS}
\label{conclusions}

In this letter, we propose the use of topologically persistent features for object recognition in indoor environments. We construct cubical complexes from binary segmentation maps of the objects. For every cubical complex, we obtain multiple filtrations using height functions in multiple directions. Persistent homology is applied to these filtrations to obtain topologically persistent features that capture the shape information of the objects. We present two different kinds of persistent features, namely, sparse PI and amplitude features, to train a fully connected recognition network. Sparse PI features achieve better recognition performance in unseen environments than features learned from ResNetV2-56 and EfficientNet-B4. Unlike end-to-end object detection methods Faster R-CNN and its state-of-the-art variant DA-FR-CNN\textbf{*}, the overall performance of our methods remains relatively unaffected on a different test environment without retraining, provided the quality of the object segmentation maps is maintained. Moreover, the sparse PI features-based method slightly outperforms DA-FR-CNN\textbf{*} in terms of recall and accuracy, making it a promising first step in achieving robust object recognition. In the future, we plan to use depth information to deal with the challenges associated with camera pose variations and to perform class-agnostic instance segmentation. Depth information, along with shape completion, might also help address the issues of incomplete segmentation maps and partial occlusion of objects in cluttered environments. We also plan to explore the use of topologically persistent features in estimating 6D object poses through few-shot deep learning.

\bibliography{references}


%%%%%%%%%%%%%%%%%%%%%%%%%%%%%%%%%%%%%%%%%%%%%%%%%%%%%%%%%%%%%%%%%%%%%%%%%%%%%%%%

%%%%%%%%%%%%%%%%%%%%%%%%%%%%%%%%%%%%%%%%%%%%%%%%%%%%%%%%%%%%%%%%%%%%%%%%%%%%%%%%

%%%%%%%%%%%%%%%%%%%%%%%%%%%%%%%%%%%%%%%%%%%%%%%%%%%%%%%%%%%%%%%%%%%%%%%%%%%%%%%%

%\section*{\textcolor{red}{ACKNOWLEDGMENT}}

%%%%%%%%%%%%%%%%%%%%%%%%%%%%%%%%%%%%%%%%%%%%%%%%%%%%%%%%%%%%%%%%%%%%%%%%%%%%%%%%

\end{document}